\pdfoutput=1

\documentclass[11pt]{article}

\usepackage[preprint]{acl}

\usepackage{times}
\usepackage{latexsym}

\usepackage[T1]{fontenc}

\usepackage[utf8]{inputenc}

\usepackage{microtype}

\usepackage{inconsolata}

\usepackage{graphicx}

\usepackage{algorithm}
\usepackage[noend]{algorithmic}
\usepackage{amsmath}
\usepackage{bm}
\usepackage{graphicx}
\usepackage{todonotes}
\usepackage{enumitem}
\usepackage{multirow}
\usepackage{booktabs}
\usepackage{amssymb}
\usepackage{titletoc}
\usepackage[most]{tcolorbox}
\usepackage{setspace}

\newcommand\DoToC{%
  \startcontents
  \printcontents{}{1}{\textbf{Contents}\vskip3pt\hrule\vskip5pt}
  \vskip3pt\hrule\vskip5pt
}
\usepackage{xspace}
\newcommand{\method}{\texttt{LRTab}\xspace}

\title{Utilizing Training Data to Improve LLM Reasoning for Tabular Understanding}

\author{Chufan Gao \\
  University of Illinois\\
  Urbana-Champaign \\
  \texttt{chufan2@illinois.edu} \\\And
  Jintai Chen \\
  The Hong Kong University \\
  of
  Science and Technology \\\And
  Jimeng Sun \\
  University of Illinois\\
  Urbana-Champaign
  }

\begin{document}
\maketitle
\begin{abstract}
Automated tabular understanding and reasoning are essential tasks for data scientists. Recently, Large language models (LLMs) have become increasingly prevalent in tabular reasoning tasks. 
Previous work focuses on (1) finetuning LLMs using labeled data or (2) Training-free prompting LLM agents using chain-of-thought (CoT). Finetuning offers dataset-specific learning at the cost of generalizability. Training-free prompting is highly generalizable but does not take full advantage of training data.
In this paper, we propose a novel prompting-based reasoning approach, \textbf{L}earn then \textbf{R}etrieve: \method, which integrates the benefits of both by retrieving relevant information learned from training data.
We first use prompting to obtain CoT responses over the \textit{training data}. For incorrect CoTs, we prompt the LLM to predict Prompt Conditions to avoid the error, learning insights from the data. We validate the effectiveness of Prompt Conditions using validation data. Finally, at inference time, we retrieve the most relevant Prompt Conditions for additional context for table understanding.
We provide comprehensive experiments on WikiTQ and Tabfact, showing that \method is interpretable, cost-efficient, and can outperform previous baselines, reaching a new SOTA in tabular reasoning.
\end{abstract}

\section{Introduction}

\begin{figure}[ht]
    \centering
    \includegraphics[width=\linewidth]{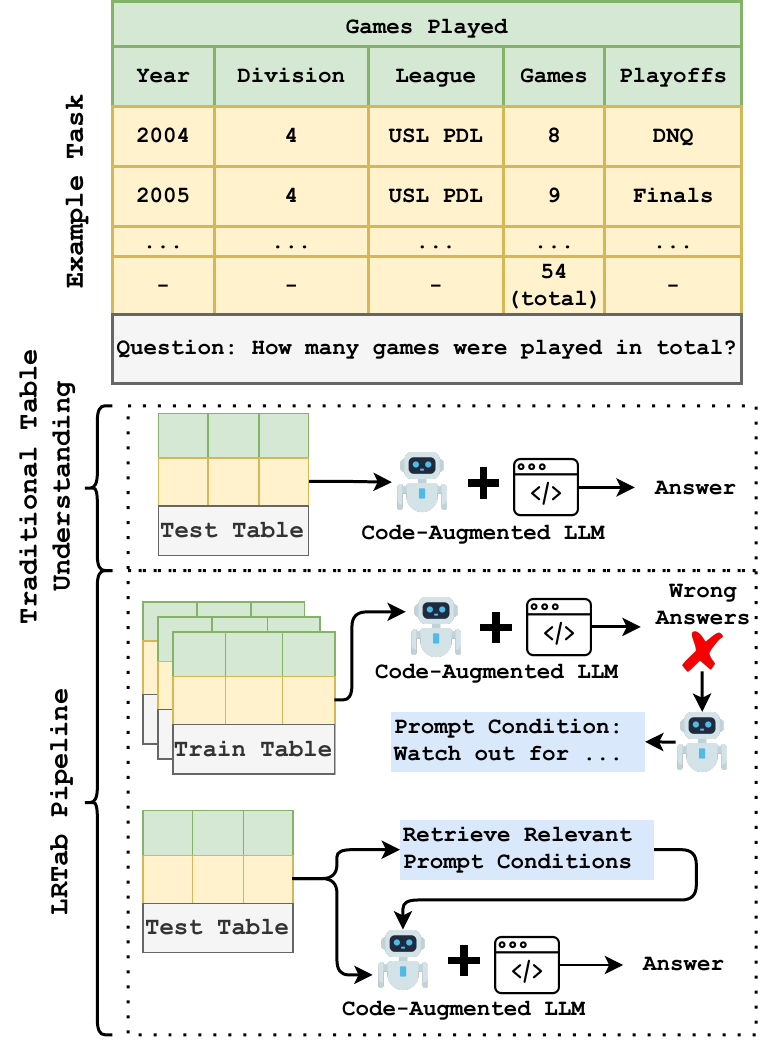}
    \caption{Overview of \method vs traditional tabular understanding methods. Traditional table understanding does not take advantage of the insights of training data, and any edge cases have to be anticipated. \method uses LLMs to generate prompt conditions to address training errors. \method also learns to retrieve and rerank these prompt conditions for inference.}
    \label{fig:enter-label}
    \vspace{-1em}
\end{figure}

Tabular data is one of the most common data types found in various business and consumer applications \cite{webtables}. Unlike text, images, or time-series data, tables are usually highly structured, easily queryable using common languages like SQL and Python, and well-defined with clear column descriptions. However, analyzing a table can still be challenging due to obscure feature names, inconsistent formatting, and sophisticated column relationships, making it difficult to derive query answers effectively. Recent advancements in large language models (LLMs) have shown potential in tabular reasoning addressing table-based fact verification~\cite{chen2019tabfact} and question answering~\cite{jin2022survey, pasupat2015compositional, nan2022fetaqa}. 

Recent efforts on LLMs for tabular reasoning primarily fall into two categories. One category involves finetuning-based approaches: i.e. training LLM embeddings/ attention modules \cite{herzig2020tapas, wang2021tuta, gu2022pasta} or training LLMs to generate better SQL queries directly \cite{eisenschlos2020understanding, liu2021tapex, jiang2022omnitab}. The other category leverages inference-only prompt engineering strategies such as Chain-of-Thought (CoT) and zero/few-shot in-context learning (ICL) ~\cite{chen2022large, cheng2022binding, ye2023large, hsieh-etal-2023-distilling, liu2023rethinking, wang2024chain} to improve the tabular reasoning capabilities of LLMs.

We believe both finetuning and inference-only prompting suffer from distinct challenges that may be mitigated by borrowing ideas from each other. Finetuning approaches are computationally costly and lack flexibility for new applications due to their reliance on task-specific labeled data. In contrast, inference-only approaches demonstrate flexibility for new applications but do not take advantage of the informative labeled data.

Recent literature has demonstrated a training-like effect in using labeled data to augment LLMs through in-context learning (ICL)~\cite{singh2023beyond, gulcehre2023reinforced, agarwal2024many} and retrieval \cite{agarwal2024many, singh2023beyond}, without training. 
However, \textit{incorrectly-reasoned} examples may reveal key knowledge gaps in LLMs, which are crucial for dataset-specific learning. Yet, current LLM prompting methods do not fully utilize or even address these \textit{incorrectly-reasoned} examples. Finetuning methods, however, utilize incorrectly predicted examples to adapt to training data via the loss function.

\begin{figure*}[ht]
    \centering
    \includegraphics[width=\linewidth]{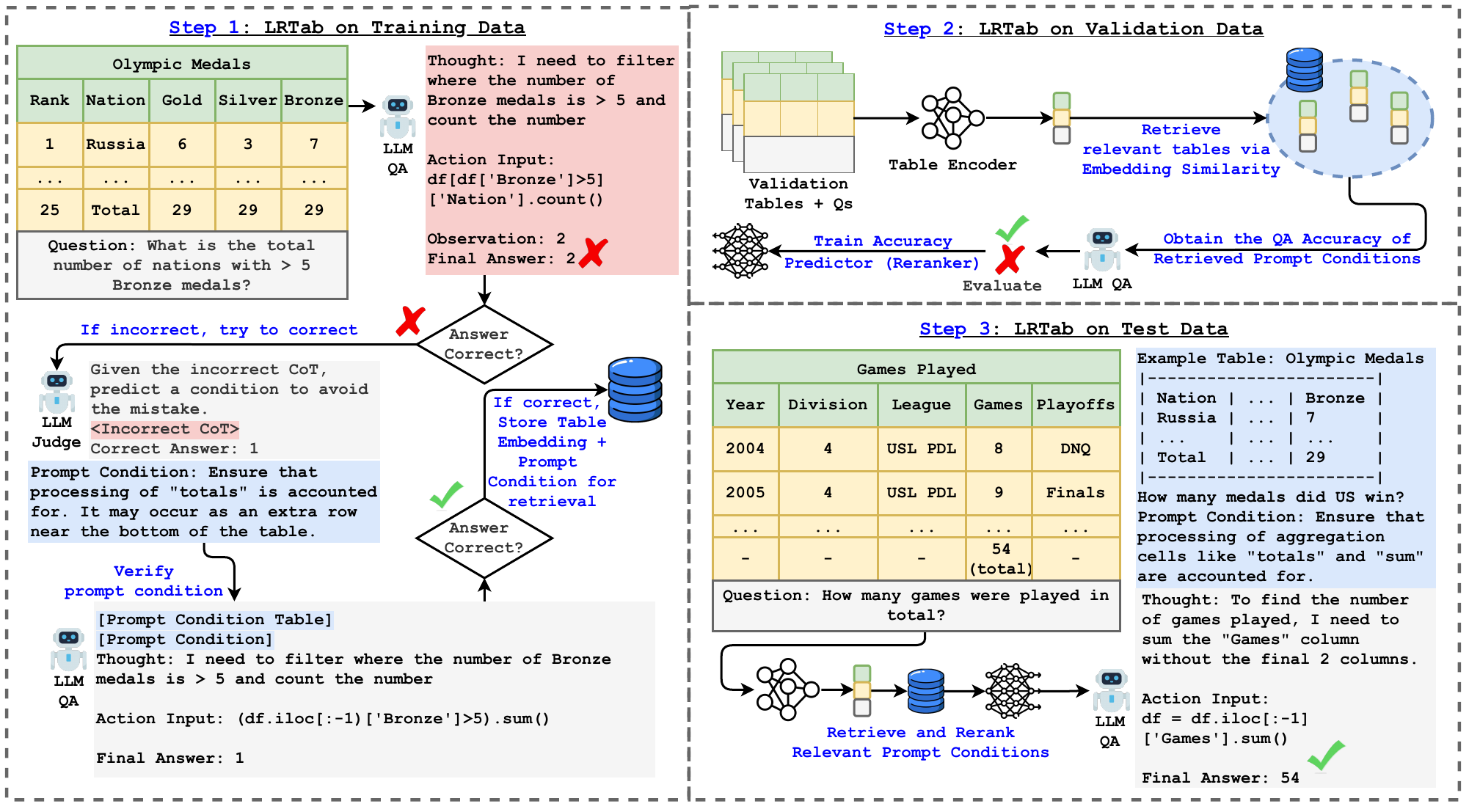}
        \caption{\method Training, Validation, and Test inference process. Step 1: LRTab on Training Data details the ``training'' phase where the model generates Chain-of-Thoughts (CoTs) for training examples. For incorrect CoTs, the LLM is prompted to generate and verify Prompt Conditions to correct the error, with both successful CoTs and conditions being stored for later retrieval. Step 2: LRTab on Validation Data outlines the process of retrieving and evaluating relevant prompt conditions using a table encoder and training a reranker to learn to retrieve relevant prompt conditions. Step 3: LRTab on Test Data illustrates the inference stage, where the LLM leverages retrieved prompt conditions and reference examples to answer new questions accurately on unseen tables.
    }
    \label{fig:method}

    \end{figure*}

In this paper, we propose \method to learn from \textit{incorrectly-reasoned} predictions on the training data to enhance the LLM's tabular reasoning capabilities at test-time. 

Our method includes ``training'', which performs tabular reasoning on labeled training data.\footnote{In this paper, ``labeled data'' indicates a (table, question, answer) triplet.}
In this step, tabular reasoning is directly performed using a code-enabled chain-of-thought process (shown in Appendix~\ref{app:prompts}).
However, contrary to previous works~\cite{agarwal2024many,gulcehre2023reinforced} that also obtain ICL examples using inference over the training set, the incorrectly-reasoned CoTs are not discarded. Instead, we prompt the LLM again to obtain \textbf{``Prompt Conditions''} that attempt to correct the errors in the incorrectly-reasoned CoT. The LLM is then prompted with the Prompt Condition to see if the error has been fixed and the predicted answer is now correct. An example prompt condition could be: \texttt{``Do not attempt to process datetimes using Python when the format is inconsistent, reason it out instead.''}


In \method, we primarily retrieve Prompt Conditions using an off-the-shelf text similarity as well as a custom crossencoder reranker on the inference table and training tables. We argue that \method is humanly interpretable and highly effective compared to finetuning approaches, and achieves better performance than inference-only approaches by combining the advantages of both.
We summarize our contribution as follows:

\begin{itemize}[leftmargin=*]
    \item The main contribution of our work, \method, lies in being able to take full advantage of labeled data for tabular reasoning, where previous inference-only LLM prompting cannot. The ``training'' step of \method allows for information extraction of insights from training data, obtaining useful, human-interpretable Prompt Conditions.
    \item At inference time, we utilize text similarity + reranking to retrieve Prompt Conditions for in-context learning (ICL). 
    Extensive experimentation in table-based fact verification and question answering demonstrates that \method achieves state-of-the-art performance on WikiTQ~\cite{pasupat2015compositional} and TabFact~\cite{chen2019tabfact}.
    \item We also analyze best practices for Table understanding, including using coding vs direct prompting, retrieval model ablations, performance by table length, as well as the number of ICL examples.
\end{itemize}
\section{Related Work}


\paragraph{Table Understanding}
Recent advancements in machine learning and data processing have led to innovative solutions for table-related QA. 
Large, pretrained LLM on multiple tables \cite{zhang2023tablellama, li2023tablegpt, jiang2022omnitab, xie2022unifiedskg} propose versatile LLMs trained to perform a variety of tasks such as reasoning, completion, QA, and more \cite{zha2023tablegpt, yang2023unitabe}. Finetuned LLMs are surprisingly good in this space, with sub-table selection and reasoning improvements \cite{zhao2022reastap, gu2022pasta, patnaik2024cabinet}. Similarly, inference-only LLMs have seen success due to their inherent reasoning abilities \cite{cheng2022binding, ye2023large, jiang2023structgpt, wang2024chain, abhyankar2024h}. Because of these advancements, we choose to use the prompting of LLMs as our backbone method for table understanding rather than fine-tuning a smaller LLM. 

\paragraph{Automated LLM Correction}
The idea of correction in LLM Agents has been recently popular \cite{agarwal2024many, singh2023beyond, gulcehre2023reinforced, shinn2024reflexion, huang2023large, feng2024thought, yuksekgonul2024textgrad}. 
The concept of ``Reinforced ICL''~\cite{agarwal2024many} evaluates the CoT rationals on labeled data and retrieves reference data in the test time. While effective, this work does not explore the idea of error case correction or adding additional Prompt Conditions. Similarly, "prethinking" on an unlabeled dataset, saving the high-confidence thoughts, and retrieving them boosts performance at inference-time for QA tasks \cite{li2023mot}. 
Huang et al. demonstrated that self-correction without ground truth does not perform well \cite{huang2023large}, which we also observed. Corrective retrieval has also been proposed \cite{yan2024corrective, asai2023self}--Asai et al. demonstrated that finetuning an LLM to learn to retrieve raw data is beneficial for QA and long-form generation \cite{asai2023self}. Self-correction is also a recently popular topic, such as the correction of SQL queries \cite{pourreza2024din, yuksekgonul2024textgrad}.
Other work also does not systematically correct incorrect CoTs \cite{shinn2024reflexion}.
However, to our knowledge, no approach has focused on the retrieval and correction of CoTs like in \method.

\section{Methodology}
For a given table-based reasoning task, we represent the given paired (table, query) as $(T, Q)$, where $T$ stands for the table and $Q$ represents a table-based question or a statement to be verified (to accommodate TabFact). 
The objective of the LLM is to predict the answer $A_{pred}$ based on the corresponding $(T, Q)$.

Figure~\ref{fig:method} demonstrates an overview of \method at inference time, and Figure~\ref{fig:ex_sc} shows \method's ``training'' step. We describe the inference stage in Section~\ref{sec:retrieval}.  We then detail the capability of \method to ``learn'' over the training data, making use of the training labels in particular. Our full algorithm is shown in Algorithm~\ref{algo:table_reasoning}.

\subsection{\method Training: Learning from Training Data}
\label{sec:training}
To obtain a set of Prompt Conditions, we first have to run regular prompting over the training dataset.

For initial tabular reasoning, we instruct LLM Agents to engage in step-by-step reasoning over the reference data. We primarily explore a flexible code-augmented setting where the LLM can \textit{choose} to run Python code on the table for further processing (a Pandas DataFrame). We explore ablations on coding in Section~\ref{sec:optional_code}.

Let $T, Q$ be the table and query pair. ($R, A_{pred}$) is the predicted CoT reasoning and predicted answer to the table QA via prompting our LLM agent $\textit{LLM}_{QA}(T, Q)$. $\textit{LLM}_{Corr}(T, Q, R, A)$ is an LLM that attempts to output a Prompt Condition $C$ to obtain the correct answer. A full algorithm is shown in Algorithm~\ref{algo:table_reasoning}.

The general tabular reasoning process is shown in Algorithm~\ref{algo:table_reasoning}. 
At each iteration, we prompt the LLM to sample its next thought in its chain of thought reasoning. 
For \method, we query the LLM for the next step, whether it be Python code, or the final answer. If it is code, we automatically execute it and update the table. 
We additionally prompt the LLM an additional time so that it can reflect on the code output in the next iteration, until either the maximum number of iterations is reached or a \texttt{"Final Answer"} as described earlier.
For the stopping criterion, we simply repeatedly append the thought to a list of new thoughts until a final answer is detected (i.e. the regex \texttt{"Final Answer:.*"} finds a match). If no answer is detected within 5 steps, we return an error.
We build on the prompt design of~\cite{liu2023rethinking} for our experiments. 

\paragraph{Correction of Incorrect CoT Rationals and Obtaining Prompt Conditions}

To motivate our reasoning for this section, our initial idea was to simply correct the LLM using the ground truth answer, and use the corrected examples as ICL in the inferece phase. \textit{Despite our best efforts, we found that any attempt to correct the LLM using the ground truth answer also referred to the answer in its reasoning}, which would not be applicable at test time. 
For example, the following would occur in the CoT reasoning: \texttt{``Given that the correct answer is X, I should ...''}.
Therefore, we split the correction method into 3 steps (shown in Figure~\ref{fig:method} Step 1: \method on Training Data):

    First, we use the LLM for inference over all training tables and queries. 
    
    Second, for the incorrect CoTs, we use the same LLM to obtain a Prompt Condition relevant to the specific mistake made in the incorrect CoT. 
    
    Third, we append the given prompt condition to the original prompt and check if it now yields the correct answer. If it does, we save the Prompt Condition (including its table and question) and CoT for retrieval.

Note that Prompt Conditions are only obtained from \textit{corrected} reference samples. Correctly answered reference samples should not need any additional conditions, as the original prompt is sufficient.

\begin{algorithm}[ht]
\small
\caption{\method Training and Validation. Test inference is the same as validation, but with an additional reranking step of the relevant tables.}
\label{algo:table_reasoning}
\textbf{Inputs:} Training set $(\textbf{T}_{train}, \textbf{Q}_{train}, \textbf{A}_{train})$; Validation set $(\textbf{T}_{test}, \textbf{Q}_{test})$.  

\textbf{Outputs:} Chain-of-thought reasoning $R$ and predicted answer $A_{pred}$ over all training and validation samples.  

\begin{algorithmic}[1] 

\STATE \textcolor{teal}{\textbf{$\vartriangleright$ Training Phase}}
\STATE Let $\textbf{C} \gets [\ ], \textbf{R} \gets [\ ], \textbf{T} \gets [\ ]$
\FOR{each training example $(T_{train}, Q_{train}, A_{train})$}
    \STATE $\textit{R}, A_{pred} \gets \textit{LLM}_{QA}(T_{train}, Q_{train})$
    
    \IF {$A_{pred} \neq A_{train}$}
        \STATE $C \gets \textit{LLM}_{Corr}(T_{train}, Q_{train}, R, A_{train})$ 
        
        \textcolor{teal}{$\vartriangleright$ Try answering again using condition $C$}\\ \STATE $R^+,A_{pred}^+ \gets \texttt{LLM}_{QA}(T_{train}, Q_{train}, C)$
        \IF {$A_{pred}^+ = A_{train}$}
            \STATE $\textbf{C} \gets \textbf{C} \cup \{\textit{C}\}$
            \STATE $\textbf{T} \gets \textbf{T} \cup \{T_{train}\}$
        \ENDIF
    \ENDIF
\ENDFOR

\STATE \textcolor{teal}{\textbf{$\vartriangleright$ Validation Phase}}

\STATE $\mathcal{D}_{val} \gets [\ ]$

\STATE $\mathbf{E}_T \gets \{\texttt{Embed}_{LLM}(T_i) \mid T_i \in \textbf{T}\}$ 
\FOR{each validation example $(T_{val}, Q_{val}, A_{val})$}
    \STATE $\mathbf{e}_{T_{val}} \gets \texttt{Embed}_{LLM}(T_{val})$ 
    
    \STATE $\textbf{T}_{\text{sim}} \gets \texttt{argmax}_k(\cos(\mathbf{e}_{T_{val}}, \mathbf{e}_{T_i}) \mid \mathbf{e}_{T_i} \in \mathbf{E}_T)$

     \textcolor{teal}{$\vartriangleright$ Get corresponding prompt conditions}\\ 
    \STATE $\Tilde{C} \gets \{C_i \mid T_i \in \textbf{T}_{\text{sim}}\}$  
    \STATE $R, A_{pred} \gets \textit{LLM}_{QA}(T_{test}, Q_{test}, \Tilde{C})$
    \STATE $\mathcal{D}_{val} \gets \mathcal{D}_{val} \cup\{(T_{val}, \textbf{T}_{\text{sim}}, A_{pred} = A_{val})\}$
\ENDFOR
\textcolor{teal}{$\vartriangleright$ Train Reranker on $\mathcal{D}_{val}$}\\ 


    

    
\end{algorithmic}
\end{algorithm}

\subsection{\method Validation: Learning to Retrieve Relevant Prompt Conditions}
\label{sec:retrieval}
This serves to update the prompt itself as a part of the finetuning analogy.

To take full advantage of the training and the insights we learned via the Prompt Conditions, we add a retrieval mechanism for ICL. Specifically, we utilize the Salesforce 400M parameter code embedding model as the encoder\footnote{\url{https://huggingface.co/https://huggingface.co/Salesforce/SFR-Embedding-Code-400M_R}} \cite{liu2024codexembed}. Retrieving relevant Prompt Conditions is important so that the in-context learning does not mislead the LLM. Retrieval is performed using text similarity on the current table vs reference tables using the following key format:
\begin{tcolorbox}{\fontfamily{lmtt} \footnotesize 
[Table in Markdown Format]

[Question / or Statement to Verify]
}\end{tcolorbox}

Retrieved Prompt Conditions are simply added to the prompt per the highlighted text in blue in Figure~\ref{fig:method} (Step 1 and Step 3).

To further enhance the quality of retrieval, we train a cross-encoder reranker on validation data. Since the validation dataset has labels, we can obtain a ``usefulness'' metric of retrieved prompt conditions. For each sample in the validation set, we retrieve $k=2$ most similar Prompt Conditions using the previously-mentioned encoder and evaluate them. Retrieved conditions are given a label of `1' if the table understanding task was the correct answer, and a label of `0' otherwise.  
We use nli-deberta-v3-large \cite{he2021debertav3} as the base crossencoder\footnote{\url{https://huggingface.co/cross-encoder/nli-deberta-v3-large}}. 

\paragraph{Inference on Unseen Data} Inference is similar to Validation, except with the additional step of using the fine-tuned crossencoder reranker. In this phase, we retrieve a higher $k=8$ relevant Prompt Conditions using the encoder, before reranking and taking the top-rated Prompt Conditions.

\section{Experiments}

\begin{table}[ht]
\centering
\caption{Dataset Statistics}
\label{tab:stats}
\resizebox{.8\linewidth}{!}{
\begin{tabular}{crrrr}
 & \multicolumn{2}{c}{WikiTQ} & \multicolumn{2}{c}{TabFact} \\ \toprule
 & Questions & Tables & Questions & Tables \\ \midrule
Train & 14,148 & 1,679 & 92,283 & 13,182 \\
Dev & 3,536 & 1,455 & 12,792 & 1,696 \\
Test & 4,344 & 421 & 2,024 & 298 \\ \bottomrule
\end{tabular}
}
\end{table} 

We evaluate \method on 2 commonly used datasets: WikiTQ~\cite{pasupat2015compositional} and TabFact~\cite{chen2019tabfact} (Table~\ref{tab:stats}). WikiTQ focuses on table-based question answering requiring complex tabular reasoning with short text span answers, while TabFact is a binary fact verification benchmark assessing the truthfulness of statements based on tables. We follow the previous works and report the performance using cleaned string matching for WikiTQ and binary prediction accuracy for TabFact respectively. 
For \method training, we use around 3000 samples of the training dataset for WikiTQ and TabFact respectively, due to budget limitations and the large size of the datasets. 

\paragraph{Baselines}
We provide an in-depth comparative analysis of the results obtained by \method against various baselines. We examine 2 distinct categories of methods. First, we compare LLM that were specifically pretrained or pretrained for table understanding and question answering, such as Unifiedskg \cite{xie2022unifiedskg}, REASTAP \cite{zhao2022reastap}, PASTA \cite{gu2022pasta}, OmniTab \cite{jiang2022omnitab}, and CABINET \cite{patnaik2024cabinet}. 

We also compare against few-shot or completely zero-shot promoting methods utilizing prompting LLMs such as BINDER \cite{cheng2022binding}, DATER \cite{ye2023large}, ReAcTable \cite{zhang2023reactable}, ProTrix \cite{wu2024protrix}, STRUCTGPT \cite{jiang2023structgpt}, Mixed Self-Consistency \cite{liu2023rethinking}, Chain-of-Table \cite{wang2024chain}, and H-STAR \cite{abhyankar2024h}. Results are reproduced whenever possible or taken from the original work otherwise. 

Note that many baselines, including Mixed Self-Consistency and H-Star, are particularly expensive to run, as they require upwards of 10 prompt calls per example. Because of this, we omit results on GPT-4o and focus on cheaper models like GPT-4o-mini.


\paragraph{\method Details}
For \method, we use GPT-4o-mini and GPT-4o. To make the tables interpretable by LLMs, we convert them into Markdown format\footnote{using the function \texttt{pandas.DataFrame.to\_markdown$(\cdot)$}}. 
While it would be interesting to perform an analysis of larger models such as GPT 4 and o1, it is prohibitively expensive to conduct experiments due to budget restrictions. To prevent token length errors, we set the number of prompt conditions to retrieve to 2 and the number of examples to retrieve to 1.

The number of CoTs and corrected examples / prompt conditions can be found in Appendex~\ref{app:sc_stats}. 

\section{Results}
\paragraph{\method vs Prompting-Based Baselines}
\begin{table}[ht]
\centering
\caption{
Accuracy comparisons of equivalent baselines vs \method. We denote (\method\ -Reranker) as the method without the reranker and \method as the full method. We reran all methods besides the results that we used from the original paper (indicated by *). Best performance is \textbf{bolded}.}
\label{tab:main}
\resizebox{\linewidth}{!}{
\begin{tabular}{llll} \toprule 
  & \begin{tabular}[c]{@{}l@{}}GPT-\\ 4o-mini\end{tabular} & \begin{tabular}[c]{@{}l@{}}Llama-3\\ 70B\end{tabular} & \begin{tabular}[c]{@{}l@{}}Gemma-3\\ 27B-IT\end{tabular} \\
\midrule
Method & \multicolumn{3}{c}{WikiTQ}\\   \midrule
COT& 64.15 & 65.33 & 68.17 \\ 
COT+Code& 67.25 & 68.42 & 69.52 \\ 
Mixed Self Consistency& 72.26 & 70.99 & 71.93 \\ 
Chain-of-Table& 65.34 & 70.76 & 70.39 \\ 
H-STAR & 74.93*& 75.76*& 75.50 \\ 
\method\ -Reranker (Ours) & 75.84 & 74.81 & 73.45 \\ 
\method (Ours) & \textbf{76.80} & \textbf{75.93} & \textbf{75.80} \\  \midrule
 & \multicolumn{3}{c}{TabFact}\\ \midrule
COT& 75.83 & 75.18 & 83.91 \\ 
COT+Code& 74.12 & 77.99 & 83.44 \\ 
Mixed Self Consistency& 86.61 & 79.49 & 87.94 \\ 
Chain-of-Table& 84.20 & 85.86 & 86.24 \\ 
H-STAR & 89.42*& 89.23*& 90.16 \\ 
\method\ -Reranker(Ours) & 89.25 & 88.66 & 89.88 \\ 
\method (Ours) & \textbf{89.74} & \textbf{89.61} & \textbf{92.24} \\ \bottomrule
\end{tabular}

}
\end{table}

Table~\ref{tab:main} presents a comparative evaluation of the best versions of \method against prior table-based QA approaches across WikiTQ and TabFact. The results demonstrate that \method achieves state-of-the-art performance, surpassing existing methods across both benchmarks.
On WikiTQ, \method with GPT-4o-mini attains 76.8\%, outperforming the next-best approach with the same base LLM, H-STAR. For TabFact, \method achieves 89.74\% with GPT-4o-mini over H-STAR, marking a smaller improvement. \method also outperforms Mixed Self-Consistency and Chain-of-Table with the same base GPT 4o-mini.
Overall, these results highlight the effectiveness of \method in table-based reasoning tasks, demonstrating that leveraging flexible reasoning strategies with advanced LLMs leads to state-of-the-art performance across both structured QA and fact verification.  We also have results on FeTaQA in Appendix~\ref{app:fetaqa}.

\paragraph{Comparison to Finetuning-Based Baselines}

\begin{table}[ht]
\caption{Test accuracy of \method and finetuning baselines. Results are copied from the original papers' most relevant and best-performing configurations (missing results are denoted with a dash ``-''). Best performance is \textbf{bolded}. \label{tab:vsfinetuning}}
\resizebox{\linewidth}{!}{
\begin{tabular}{llll} \toprule
\textbf{Approach}& \textbf{Base Model}& \textbf{WikiTQ}& \textbf{TabFact}\\ \midrule
Unifiedskg& T5 3B& 49.29& 83.68\\
REASTAP& BART-Large& 58.6& 80.1\\
PASTA& DeBERTaV3& -& 85.60\\
OmniTab& BART-Large& 62.80& -\\
CABINET& BART-Large& 69.10& -\\ \midrule
 \method (Ours)& LLama-3-70B& 75.93&89.61\\
 \method (Ours)& Gemma-3-27B-IT& 75.80&92.24\\
\method (Ours)& GPT 4o-mini& 76.80& 89.74\\
\method (Ours)& GPT 4o& \textbf{80.02}& \textbf{93.38}\\ \bottomrule
\end{tabular}
}
\end{table}

In this section, we overview \method in comparison to other finetuning-based table understanding baselines--Unifiedskg \cite{xie2022unifiedskg}, REASTAP \cite{zhao2022reastap}, PASTA \cite{gu2022pasta}, OmniTab \cite{jiang2022omnitab}, and CABINET \cite{patnaik2024cabinet}. 
From Table~\ref{tab:vsfinetuning}, we see that \method outperforms the other baselines. Additionally, since we were unable to run GPT-4o for all baselines in Table~\ref{tab:main}, we chose to show the results here as an example of our model's best performance. It is important to note that both WikiTQ as well as TabFact are very general in their domain. This is good for general-purpose LLMs like GPT 4o, which have broad knowledge. Finetuning allows much smaller LLMs, such as DeBERTaV3, to perform quite well, as seen in its TabFact performance. Still, \method excels considering its ease of use and its good performance.

Note that due to the high cost of inference over the entire dataset, we evaluate the rest of the experiments on a subset of 500 test samples. Additional comparisons are shown in Appendix~\ref{app:addition_experiments}.

\paragraph{Retrieving Prompt Conditions vs Examples}
Correctly-answered CoT samples are useful for in-context learning \cite{agarwal2024many}. Intuitively, this acts like a grounding mechanism for the LLM generation, as it has an example of what a correctly reasoned CoT should look like. Still, this comes with a lot of downsides, as these examples may be quite long in practice, and longer prompt length is a detriment to performance. Table~\ref{tab:collab_diffs} examines the effects of retrieving prompt conditions, chain-of-thought (CoT) examples, and combining both.

\begin{table}[ht]
\centering
\caption{Results on WikiTQ and TabFact for all configurations of \method (different additions to the inference prompt).}
\label{tab:collab_diffs}
\resizebox{\linewidth}{!}{
\setlength{\tabcolsep}{0.3em} 
\Large
\begin{tabular}{lcccc} \toprule
 & \multicolumn{2}{c}{\textbf{WikiTQ}} & \multicolumn{2}{c}{\textbf{TabFact}} \\ 
Prompt Additions & GPT 4o-mini & GPT 4o & GPT 4o-mini & GPT 4o \\ \midrule
 None & 73.86 & 76.25 & 87.02 &92.20 \\ 
W/ Prompt Cond. & 75.84 & 77.95 & 89.25 & 93.34 \\
W/ CoT  & 74.39 & 78.12 & 83.37 & 93.21 \\
W/ Both & 72.50 & 78.27 & 78.34 & 93.75 \\ \bottomrule
\end{tabular}
}
\end{table}

The results reveal that while prompt augmentation strategies can enhance performance, their effectiveness depends on both the model scale and task. 

For WikiTQ, retrieving prompt conditions consistently improves accuracy over the baseline, particularly for the smaller model, suggesting that access to relevant contextual cues helps with structured queries. CoT examples, however, show mixed results—while they benefit the stronger model, they provide limited gains for the smaller one, likely due to increased prompt complexity. Interestingly, combining both strategies does not yield additive benefits and even leads to a performance drop for the smaller model, indicating that longer prompts may introduce noise or exceed optimal context windows.
Similarly, in TabFact, prompt conditions by themselves improve performance across both models, but CoT examples enhance performance for the stronger model but degrade it for the smaller one. 

The best results are achieved by combining both strategies in the stronger model, whereas the smaller model suffers from excessive prompt length.
These findings show that scaling prompt complexity with model capacity is highly important for slightly less powerful LLMs.

\textit{For the general case, retrieving prompt conditions may be sufficient to improve model performance.}



\paragraph{Coding vs No Coding}
\label{sec:optional_code}
Since weight initiation is highly important in training any neural network, we attempt to find the best possible table understanding backbone. In many related works \cite{liu2023rethinking, abhyankar2024h}, the agent is allowed to use programming languages such as Python or SQL to process the table. Inspired by this, we examine all 3 ways of allowing LLMs to interact with a Python interpreter--never using code, always using code, or letting the model decide when to use code.

\begin{table}[ht]
\caption{Performance on Test Set based on choice of table QA LLM agent. \label{tab:abl_qa_agent}}
\resizebox{\linewidth}{!}{
\Huge
\begin{tabular}{lcccc} \toprule
 & \multicolumn{2}{c}{WikiTQ} & \multicolumn{2}{c}{TabFact} \\
Table QA Method & GPT 4o-mini & GPT 4o & GPT 4o-mini & GPT 4o \\ \midrule
Code Agent & 71.02 & 73.13 & 84.14 & 89.31 \\
Direct Prompting & 70.84 & 73.34 & 84.27 & 89.51 \\
Flexible (Ours) & 73.86 & 76.25 & 87.02 & 92.20 \\ \bottomrule
\end{tabular}
}
\end{table}

The results in Table~\ref{tab:abl_qa_agent} indicate that a flexible prompting approach, which dynamically selects between direct inference and code execution, consistently outperforms both static prompting strategies across WikiTQ and TabFact benchmarks. Our finding aligns with similar work in steering LLMs \citet{chen2024steering}. Notably, the flexible method achieves the highest accuracy across all settings, with improvements of up to 3 points over direct prompting and 2.8 points over the code agent. 

For WikiTQ, the performance of direct prompting and the code agent is comparable, with a minor gain when employing code for GPT-4o-mini but a slight degradation for GPT-4o. This suggests that for structured QA, explicit code execution does not always enhance model performance. However, allowing the model to adaptively choose its reasoning strategy yields significant improvements, particularly for GPT-4o (76.25\%).

A similar trend is observed in TabFact, where direct prompting slightly outperforms the code agent. The flexible approach, however, leads to the most substantial gains, improving accuracy by 2.75 and 2.69 points for GPT-4o-mini and GPT-4o, respectively. These findings suggest that adaptively integrating symbolic reasoning via code with direct LLM inference leads to more robust table understanding and fact verification, demonstrating the efficacy of hybrid reasoning paradigms. 

\textit{In short, LLMs perform better when given the option to use code or not. We use the flexible approach for all of the experiments in this paper.
}
\paragraph{Text Encoder vs Random Retrieval}

\begin{table}[ht]
\centering
\caption{Comparison of Prompt Condition Retrieval Methods. Similarity-based indicates table text similarity.}
\label{tab:pr_retrieve}
\resizebox{\linewidth}{!}{
\setlength{\tabcolsep}{0.3em} 
\Large
\begin{tabular}{lllll} \toprule
 & \multicolumn{2}{c}{WikiTQ} & \multicolumn{2}{c}{TabFact} \\ 
Prompt Conditions & GPT 4o-mini & GPT 4o & GPT 4o-mini & GPT 4o \\ \midrule
None & 73.86 & 76.25 & 87.02 & 92.20 \\
Random & 73.24 & 76.53 & 87.15 & 92.17 \\
Similarity-based & 75.84 & 77.95 & 89.25 & 93.34 \\ \bottomrule
\end{tabular}
}
\end{table}
Table~\ref{tab:pr_retrieve} presents a sanity check on the effectiveness of retrieving relevant prompt conditions by comparing random retrieval, no retrieval, and similarity-based retrieval. The results confirm that retrieving semantically relevant conditions improves performance, while random retrieval provides no meaningful benefit over the baseline.
Random retrieval performs comparably to or worse than no retrieval, suggesting that irrelevant context may introduce noise rather than aid reasoning. Interestingly, while similarity-based retrieval yields substantial improvements, the gains are more pronounced for the smaller model. This suggests that weaker models benefit more from explicit guidance, whereas stronger models like GPT-4o already exhibit robust reasoning capabilities, reducing the marginal benefit of retrieval.


\paragraph{Encoders and Rerankers}
\begin{figure}[ht]
    \centering
    \includegraphics[width=\linewidth]{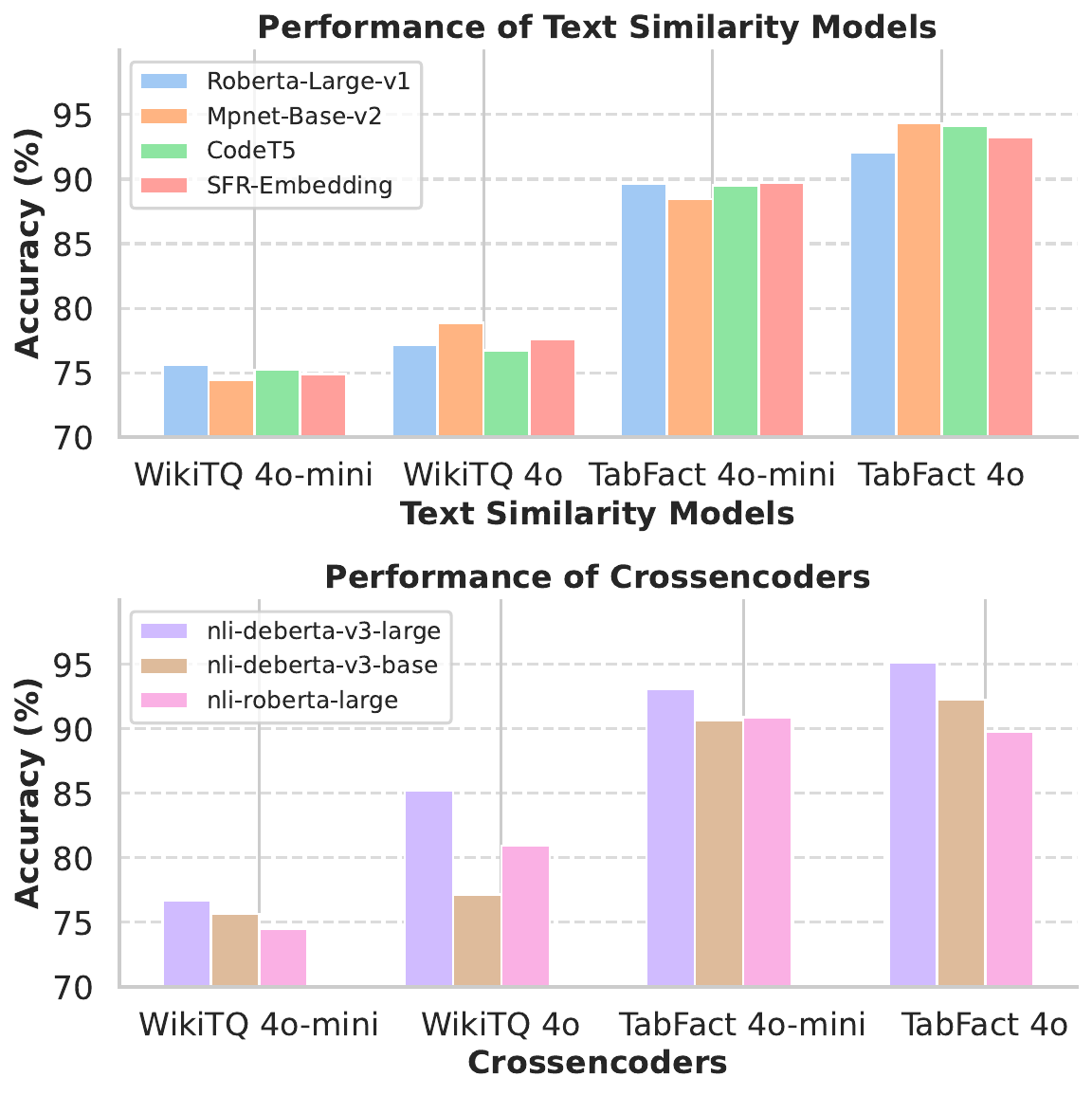}
    \caption{Ablation on varying text similarity models for
retrieving prompt conditions}
    \label{fig:text_sim}
\end{figure}
Table~\ref{fig:text_sim} shows that retrieval model choice has a minimal effect on final performance, with all models yielding nearly identical results across WikiTQ and TabFact.
Across both datasets, SFR-Embedding achieves the highest overall accuracy, but the differences between models are marginal (<0.1\%  in most cases). 

This suggests that as long as a sufficiently strong embedding model is used, the specific choice of similarity model does not significantly impact performance. However, this is not the case for cross-encoders, as larger cro ss-encoders seem to perform better than smaller ones.

\paragraph{\method Retrieval Scaling Ablations}
\label{sec:low_data}
\begin{figure}[ht]
    \centering
    \includegraphics[width=\linewidth]{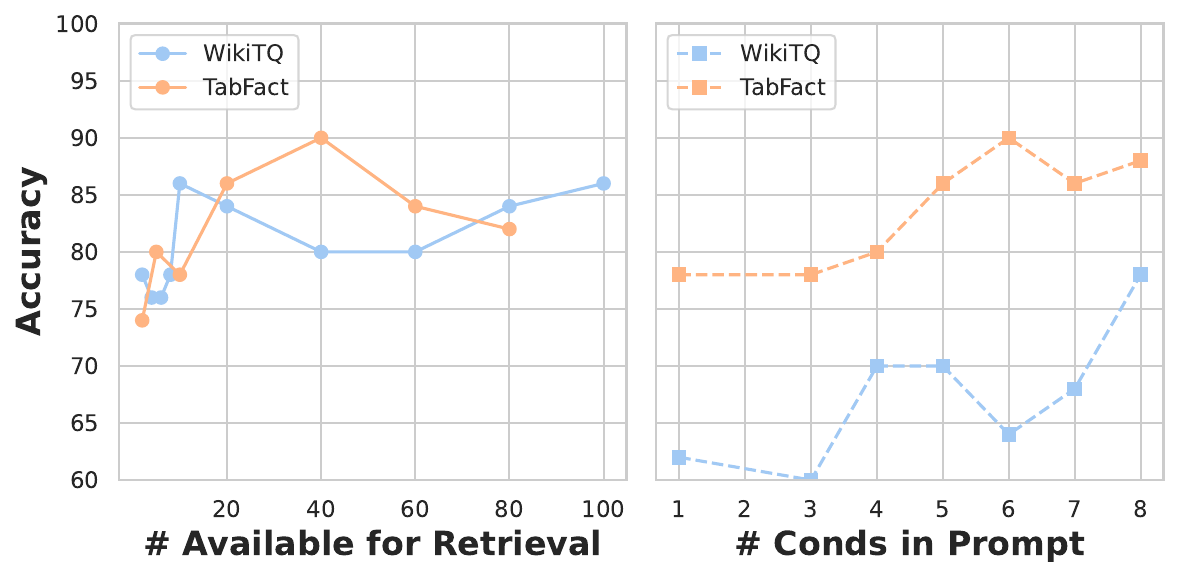}
    \caption{Scaling experiments. \textbf{Left}: The number of available prompt conditions to retrieve over. \textbf{Right}: The number of prompt conditions we retrieve per datapoint and add to the prompt.}
    \label{fig:scaling}
\end{figure}

We demonstrate experiments on scaling varying aspects of the retrieval mechanism, shown in Figure~\ref{fig:scaling}. We see that there is a positive correlation between the number of retrievable prompt conditions with performance. A similar correlation exists for the number of conditions in the prompt. Although we chose to keep the number of prompt conditions and training data small due to budget constraints, this trend is especially exciting for future work.

\paragraph{Performance by Table Length}
\begin{table}[ht]
\centering
\caption{Accuracy of \method and other baselines on varying table token lengths. Short is $<2,000$; Medium is $\leq2,000$ and $<4,000$; Long is $\geq 4,000$. }
\label{tab:tablen}
\resizebox{.8\linewidth}{!}{
\begin{tabular}{llll} \toprule
Model & Short & Medium & Long \\ \midrule
BINDER & 56.54 & 26.13 & 6.41 \\
DATER & 62.50 & 42.34 & 34.62 \\
Chain-of-Table* 4o-mini & 66.51 & 45.94 & 36.32 \\
Chain-of-Table* 4o & 70.02 & 53.12 & 42.24 \\
Chain-of-Table (PaLM) & 68.13 & 52.25 & 44.87 \\ \midrule
\method (llama-3-70B) & 73.84 & 68.75 & 59.94 \\
\method (gemma-27b) & 77.83 & 65.62 & 57.18 \\
\method (4o-mini) & 78.24 & 72.01 & 62.59 \\
\method (4o) & 82.09 & 76.41 & 66.67 \\ \bottomrule
\end{tabular}

}
\end{table} 
Table~\ref{tab:tablen} presents a comparative analysis of different methods applied to tables of varying token lengths. Similar to previous baselines, \method variations show clear degradation as table size increases. Overall, we observe superior performance in all categories, particularly in handling medium and long tables.


\section{Conclusion}

\method presents a significant step towards more effective and interpretable tabular reasoning by uniquely harnessing the full potential of labeled training data through a prompting-based framework. It learns from both successes and failures in the training phase to create actionable, human-readable guidance in the form of Prompt Conditions. These conditions, retrieved at inference time, help the LLM avoid past mistakes and reason more accurately, leading to state-of-the-art performance on the WikiTQ and TabFact datasets across different table lengths. 

Looking ahead, LRTab opens up several exciting avenues for future research. There is considerable potential in extending this methodology to accommodate very long-context LLMs and tackling the complexities of extremely long or multi-table question answering. Additionally, developing techniques for few-shot domain adaptation could mitigate the current reliance on substantial initial training data, making LRTab more versatile in low-data regimes. 

\newpage
\section*{Limitations}
\method is a good first step into fully utilizing the labels and the incorrect CoT rationales, but it is not without limitations. 
Prompt Condition retrieval may be generally helpful for tabular reasoning for tasks that are from the same domain, but one must take care to verify this, as there is a concern of "overfitting" the training data otherwise.
We were unable to run certain text-gradient baselines, such as TextGrad, as applying it to multi-turn table QA reasoning pipelines would require extensive engineering to support branching reasoning paths, as our model may be prompted up to 5 times per sample. However, this is particularly exciting, so we leave this for future work.
Furthermore, future work should seek to evaluate \method on very large LLMs with an extremely long context or long table QA to test the limits of retrieval.
Furthermore, in truly new applications, data collection and annotation may be difficult, limiting \method in such cases. Future work may explore few-shot domain adaptations to low-data regimes.
\bibliography{custom}

\begin{thebibliography}{42}
\providecommand{\natexlab}[1]{#1}

\bibitem[{Abhyankar et~al.(2024)Abhyankar, Gupta, Roth, and Reddy}]{abhyankar2024h}
Nikhil Abhyankar, Vivek Gupta, Dan Roth, and Chandan~K Reddy. 2024.
\newblock H-star: Llm-driven hybrid sql-text adaptive reasoning on tables.
\newblock \emph{arXiv preprint arXiv:2407.05952}.

\bibitem[{Agarwal et~al.(2024)Agarwal, Singh, Zhang, Bohnet, Chan, Anand, Abbas, Nova, Co-Reyes, Chu et~al.}]{agarwal2024many}
Rishabh Agarwal, Avi Singh, Lei~M Zhang, Bernd Bohnet, Stephanie Chan, Ankesh Anand, Zaheer Abbas, Azade Nova, John~D Co-Reyes, Eric Chu, et~al. 2024.
\newblock Many-shot in-context learning.
\newblock \emph{arXiv preprint arXiv:2404.11018}.

\bibitem[{Asai et~al.(2023)Asai, Wu, Wang, Sil, and Hajishirzi}]{asai2023self}
Akari Asai, Zeqiu Wu, Yizhong Wang, Avirup Sil, and Hannaneh Hajishirzi. 2023.
\newblock Self-rag: Learning to retrieve, generate, and critique through self-reflection.
\newblock \emph{arXiv preprint arXiv:2310.11511}.

\bibitem[{Cafarella et~al.(2008)Cafarella, Halevy, Wang, Wu, and Zhang}]{webtables}
Michael~J. Cafarella, Alon Halevy, Daisy~Zhe Wang, Eugene Wu, and Yang Zhang. 2008.
\newblock \href {https://doi.org/10.14778/1453856.1453916} {Webtables: Exploring the power of tables on the web}.
\newblock \emph{Proc. VLDB Endow.}, 1(1):538–549.

\bibitem[{Chen(2023)}]{chen2022large}
Wenhu Chen. 2023.
\newblock \href {https://doi.org/10.18653/v1/2023.findings-eacl.83} {Large language models are few(1)-shot table reasoners}.
\newblock In \emph{Findings of the Association for Computational Linguistics: EACL 2023}, pages 1120--1130, Dubrovnik, Croatia. Association for Computational Linguistics.

\bibitem[{Chen et~al.(2019)Chen, Wang, Chen, Zhang, Wang, Li, Zhou, and Wang}]{chen2019tabfact}
Wenhu Chen, Hongmin Wang, Jianshu Chen, Yunkai Zhang, Hong Wang, Shiyang Li, Xiyou Zhou, and William~Yang Wang. 2019.
\newblock Tabfact: A large-scale dataset for table-based fact verification.
\newblock In \emph{International Conference on Learning Representations}.

\bibitem[{Chen et~al.(2024)Chen, Jhamtani, Sharma, Fan, and Wang}]{chen2024steering}
Yongchao Chen, Harsh Jhamtani, Srinagesh Sharma, Chuchu Fan, and Chi Wang. 2024.
\newblock Steering large language models between code execution and textual reasoning.
\newblock \emph{arXiv preprint arXiv:2410.03524}.

\bibitem[{Cheng et~al.(2022)Cheng, Xie, Shi, Li, Nadkarni, Hu, Xiong, Radev, Ostendorf, Zettlemoyer et~al.}]{cheng2022binding}
Zhoujun Cheng, Tianbao Xie, Peng Shi, Chengzu Li, Rahul Nadkarni, Yushi Hu, Caiming Xiong, Dragomir Radev, Mari Ostendorf, Luke Zettlemoyer, et~al. 2022.
\newblock Binding language models in symbolic languages.
\newblock In \emph{International Conference on Learning Representations}.

\bibitem[{Eisenschlos et~al.(2020)Eisenschlos, Krichene, and M{\"u}ller}]{eisenschlos2020understanding}
Julian Eisenschlos, Syrine Krichene, and Thomas M{\"u}ller. 2020.
\newblock \href {https://doi.org/10.18653/v1/2020.findings-emnlp.27} {Understanding tables with intermediate pre-training}.
\newblock In \emph{Findings of the Association for Computational Linguistics: EMNLP 2020}, pages 281--296, Online. Association for Computational Linguistics.

\bibitem[{Feng et~al.(2024)Feng, Han, Lin, Liu, and You}]{feng2024thought}
Tao Feng, Pengrui Han, Guanyu Lin, Ge~Liu, and Jiaxuan You. 2024.
\newblock Thought-retriever: Don’t just retrieve raw data, retrieve thoughts.
\newblock In \emph{ICLR 2024 Workshop: How Far Are We From AGI}.

\bibitem[{Gu et~al.(2022)Gu, Fan, Tang, Nakov, Zhao, and Du}]{gu2022pasta}
Zihui Gu, Ju~Fan, Nan Tang, Preslav Nakov, Xiaoman Zhao, and Xiaoyong Du. 2022.
\newblock \href {https://doi.org/10.18653/v1/2022.emnlp-main.331} {{PASTA}: Table-operations aware fact verification via sentence-table cloze pre-training}.
\newblock In \emph{Proceedings of the 2022 Conference on Empirical Methods in Natural Language Processing}, pages 4971--4983, Abu Dhabi, United Arab Emirates. Association for Computational Linguistics.

\bibitem[{Gulcehre et~al.(2023)Gulcehre, Paine, Srinivasan, Konyushkova, Weerts, Sharma, Siddhant, Ahern, Wang, Gu et~al.}]{gulcehre2023reinforced}
Caglar Gulcehre, Tom~Le Paine, Srivatsan Srinivasan, Ksenia Konyushkova, Lotte Weerts, Abhishek Sharma, Aditya Siddhant, Alex Ahern, Miaosen Wang, Chenjie Gu, et~al. 2023.
\newblock Reinforced self-training (rest) for language modeling.
\newblock \emph{arXiv preprint arXiv:2308.08998}.

\bibitem[{He et~al.(2021)He, Gao, and Chen}]{he2021debertav3}
Pengcheng He, Jianfeng Gao, and Weizhu Chen. 2021.
\newblock \href {https://arxiv.org/abs/2111.09543} {Debertav3: Improving deberta using electra-style pre-training with gradient-disentangled embedding sharing}.
\newblock \emph{Preprint}, arXiv:2111.09543.

\bibitem[{Herzig et~al.(2020)Herzig, Nowak, M{\"u}ller, Piccinno, and Eisenschlos}]{herzig2020tapas}
Jonathan Herzig, Pawel~Krzysztof Nowak, Thomas M{\"u}ller, Francesco Piccinno, and Julian Eisenschlos. 2020.
\newblock \href {https://doi.org/10.18653/v1/2020.acl-main.398} {{T}a{P}as: Weakly supervised table parsing via pre-training}.
\newblock In \emph{Proceedings of the 58th Annual Meeting of the Association for Computational Linguistics}, pages 4320--4333, Online. Association for Computational Linguistics.

\bibitem[{Hsieh et~al.(2023)Hsieh, Li, Yeh, Nakhost, Fujii, Ratner, Krishna, Lee, and Pfister}]{hsieh-etal-2023-distilling}
Cheng-Yu Hsieh, Chun-Liang Li, Chih-kuan Yeh, Hootan Nakhost, Yasuhisa Fujii, Alex Ratner, Ranjay Krishna, Chen-Yu Lee, and Tomas Pfister. 2023.
\newblock Distilling step-by-step! outperforming larger language models with less training data and smaller model sizes.
\newblock In \emph{Findings of the Association for Computational Linguistics: ACL 2023}. Association for Computational Linguistics.

\bibitem[{Huang et~al.(2023)Huang, Chen, Mishra, Zheng, Yu, Song, and Zhou}]{huang2023large}
Jie Huang, Xinyun Chen, Swaroop Mishra, Huaixiu~Steven Zheng, Adams~Wei Yu, Xinying Song, and Denny Zhou. 2023.
\newblock Large language models cannot self-correct reasoning yet.
\newblock \emph{arXiv preprint arXiv:2310.01798}.

\bibitem[{Jiang et~al.(2023)Jiang, Zhou, Dong, Ye, Zhao, and Wen}]{jiang2023structgpt}
Jinhao Jiang, Kun Zhou, Zican Dong, Keming Ye, Wayne~Xin Zhao, and Ji-Rong Wen. 2023.
\newblock Structgpt: A general framework for large language model to reason over structured data.
\newblock \emph{arXiv preprint arXiv:2305.09645}.

\bibitem[{Jiang et~al.(2022)Jiang, Mao, He, Neubig, and Chen}]{jiang2022omnitab}
Zhengbao Jiang, Yi~Mao, Pengcheng He, Graham Neubig, and Weizhu Chen. 2022.
\newblock \href {https://doi.org/10.18653/v1/2022.naacl-main.68} {{O}mni{T}ab: Pretraining with natural and synthetic data for few-shot table-based question answering}.
\newblock In \emph{Proceedings of the 2022 Conference of the North American Chapter of the Association for Computational Linguistics: Human Language Technologies}, pages 932--942, Seattle, United States. Association for Computational Linguistics.

\bibitem[{Jin et~al.(2022)Jin, Siebert, Li, and Chen}]{jin2022survey}
Nengzheng Jin, Joanna Siebert, Dongfang Li, and Qingcai Chen. 2022.
\newblock A survey on table question answering: recent advances.
\newblock In \emph{China Conference on Knowledge Graph and Semantic Computing}, pages 174--186. Springer.

\bibitem[{Li et~al.(2023)Li, He, Yashar, Cui, Ge, Zhang, Fainman, Zhang, and Chaudhuri}]{li2023tablegpt}
Peng Li, Yeye He, Dror Yashar, Weiwei Cui, Song Ge, Haidong Zhang, Danielle~Rifinski Fainman, Dongmei Zhang, and Surajit Chaudhuri. 2023.
\newblock Table-gpt: Table-tuned gpt for diverse table tasks.
\newblock \emph{arXiv preprint arXiv:2310.09263}.

\bibitem[{Li and Qiu(2023)}]{li2023mot}
Xiaonan Li and Xipeng Qiu. 2023.
\newblock Mot: Memory-of-thought enables chatgpt to self-improve.
\newblock In \emph{Proceedings of the 2023 Conference on Empirical Methods in Natural Language Processing}, pages 6354--6374.

\bibitem[{Liu et~al.(2021)Liu, Chen, Guo, Ziyadi, Lin, Chen, and Lou}]{liu2021tapex}
Qian Liu, Bei Chen, Jiaqi Guo, Morteza Ziyadi, Zeqi Lin, Weizhu Chen, and Jian-Guang Lou. 2021.
\newblock {TAPEX}: Table pre-training via learning a neural sql executor.
\newblock In \emph{International Conference on Learning Representations}.

\bibitem[{Liu et~al.(2023)Liu, Wang, and Chen}]{liu2023rethinking}
Tianyang Liu, Fei Wang, and Muhao Chen. 2023.
\newblock Rethinking tabular data understanding with large language models.
\newblock \emph{arXiv preprint arXiv:2312.16702}.

\bibitem[{Liu et~al.(2024)Liu, Meng, Jot, Savarese, Xiong, Zhou, and Yavuz}]{liu2024codexembed}
Ye~Liu, Rui Meng, Shafiq Jot, Silvio Savarese, Caiming Xiong, Yingbo Zhou, and Semih Yavuz. 2024.
\newblock Codexembed: A generalist embedding model family for multiligual and multi-task code retrieval.
\newblock \emph{arXiv preprint arXiv:2411.12644}.

\bibitem[{Nan et~al.(2022)Nan, Hsieh, Mao, Lin, Verma, Zhang, Kry{\'s}ci{\'n}ski, Schoelkopf, Kong, Tang, Mutuma, Rosand, Trindade, Bandaru, Cunningham, Xiong, Radev, and Radev}]{nan2022fetaqa}
Linyong Nan, Chiachun Hsieh, Ziming Mao, Xi~Victoria Lin, Neha Verma, Rui Zhang, Wojciech Kry{\'s}ci{\'n}ski, Hailey Schoelkopf, Riley Kong, Xiangru Tang, Mutethia Mutuma, Ben Rosand, Isabel Trindade, Renusree Bandaru, Jacob Cunningham, Caiming Xiong, Dragomir Radev, and Dragomir Radev. 2022.
\newblock \href {https://doi.org/10.1162/tacl\_a\_00446} {{F}e{T}a{QA}: Free-form table question answering}.
\newblock \emph{Transactions of the Association for Computational Linguistics}, 10:35--49.

\bibitem[{Pasupat and Liang(2015)}]{pasupat2015compositional}
Panupong Pasupat and Percy Liang. 2015.
\newblock \href {https://doi.org/10.3115/v1/P15-1142} {Compositional semantic parsing on semi-structured tables}.
\newblock In \emph{Proceedings of the 53rd Annual Meeting of the Association for Computational Linguistics and the 7th International Joint Conference on Natural Language Processing (Volume 1: Long Papers)}, pages 1470--1480, Beijing, China. Association for Computational Linguistics.

\bibitem[{Patnaik et~al.(2024)Patnaik, Changwal, Aggarwal, Bhatia, Kumar, and Krishnamurthy}]{patnaik2024cabinet}
Sohan Patnaik, Heril Changwal, Milan Aggarwal, Sumita Bhatia, Yaman Kumar, and Balaji Krishnamurthy. 2024.
\newblock Cabinet: Content relevance based noise reduction for table question answering.
\newblock \emph{arXiv preprint arXiv:2402.01155}.

\bibitem[{Pourreza and Rafiei(2024)}]{pourreza2024din}
Mohammadreza Pourreza and Davood Rafiei. 2024.
\newblock Din-sql: Decomposed in-context learning of text-to-sql with self-correction.
\newblock \emph{Advances in Neural Information Processing Systems}, 36.

\bibitem[{Shinn et~al.(2024)Shinn, Cassano, Gopinath, Narasimhan, and Yao}]{shinn2024reflexion}
Noah Shinn, Federico Cassano, Ashwin Gopinath, Karthik Narasimhan, and Shunyu Yao. 2024.
\newblock Reflexion: Language agents with verbal reinforcement learning.
\newblock \emph{Advances in Neural Information Processing Systems}, 36.

\bibitem[{Singh et~al.(2023)Singh, Co-Reyes, Agarwal, Anand, Patil, Liu, Harrison, Lee, Xu, Parisi et~al.}]{singh2023beyond}
Avi Singh, John~D Co-Reyes, Rishabh Agarwal, Ankesh Anand, Piyush Patil, Peter~J Liu, James Harrison, Jaehoon Lee, Kelvin Xu, Aaron Parisi, et~al. 2023.
\newblock Beyond human data: Scaling self-training for problem-solving with language models.
\newblock \emph{arXiv preprint arXiv:2312.06585}.

\bibitem[{Wang et~al.(2021)Wang, Dong, Jia, Li, Fu, Han, and Zhang}]{wang2021tuta}
Zhiruo Wang, Haoyu Dong, Ran Jia, Jia Li, Zhiyi Fu, Shi Han, and Dongmei Zhang. 2021.
\newblock {TUTA}: Tree-based transformers for generally structured table pre-training.
\newblock In \emph{Proceedings of the 27th ACM SIGKDD Conference on Knowledge Discovery \& Data Mining}, pages 1780--1790.

\bibitem[{Wang et~al.(2024)Wang, Zhang, Li, Eisenschlos, Perot, Wang, Miculicich, Fujii, Shang, Lee et~al.}]{wang2024chain}
Zilong Wang, Hao Zhang, Chun-Liang Li, Julian~Martin Eisenschlos, Vincent Perot, Zifeng Wang, Lesly Miculicich, Yasuhisa Fujii, Jingbo Shang, Chen-Yu Lee, et~al. 2024.
\newblock Chain-of-table: Evolving tables in the reasoning chain for table understanding.
\newblock \emph{arXiv preprint arXiv:2401.04398}.

\bibitem[{Wu and Feng(2024)}]{wu2024protrix}
Zirui Wu and Yansong Feng. 2024.
\newblock Protrix: building models for planning and reasoning over tables with sentence context.
\newblock \emph{arXiv preprint arXiv:2403.02177}.

\bibitem[{Xie et~al.(2022)Xie, Wu, Shi, Zhong, Scholak, Yasunaga, Wu, Zhong, Yin, Wang et~al.}]{xie2022unifiedskg}
Tianbao Xie, Chen~Henry Wu, Peng Shi, Ruiqi Zhong, Torsten Scholak, Michihiro Yasunaga, Chien-Sheng Wu, Ming Zhong, Pengcheng Yin, Sida~I Wang, et~al. 2022.
\newblock Unifiedskg: Unifying and multi-tasking structured knowledge grounding with text-to-text language models.
\newblock \emph{arXiv preprint arXiv:2201.05966}.

\bibitem[{Yan et~al.(2024)Yan, Gu, Zhu, and Ling}]{yan2024corrective}
Shi-Qi Yan, Jia-Chen Gu, Yun Zhu, and Zhen-Hua Ling. 2024.
\newblock Corrective retrieval augmented generation.
\newblock \emph{arXiv preprint arXiv:2401.15884}.

\bibitem[{Yang et~al.(2023)Yang, Wang, Liu, Wu, and Liu}]{yang2023unitabe}
Yazheng Yang, Yuqi Wang, Guang Liu, Ledell Wu, and Qi~Liu. 2023.
\newblock Unitabe: Pretraining a unified tabular encoder for heterogeneous tabular data.
\newblock \emph{arXiv preprint arXiv:2307.09249}.

\bibitem[{Ye et~al.(2023)Ye, Hui, Yang, Li, Huang, and Li}]{ye2023large}
Yunhu Ye, Binyuan Hui, Min Yang, Binhua Li, Fei Huang, and Yongbin Li. 2023.
\newblock Large language models are versatile decomposers: Decompose evidence and questions for table-based reasoning.
\newblock \emph{arXiv preprint arXiv:2301.13808}.

\bibitem[{Yuksekgonul et~al.(2024)Yuksekgonul, Bianchi, Boen, Liu, Huang, Guestrin, and Zou}]{yuksekgonul2024textgrad}
Mert Yuksekgonul, Federico Bianchi, Joseph Boen, Sheng Liu, Zhi Huang, Carlos Guestrin, and James Zou. 2024.
\newblock Textgrad: Automatic" differentiation" via text.
\newblock \emph{arXiv preprint arXiv:2406.07496}.

\bibitem[{Zha et~al.(2023)Zha, Zhou, Li, Wang, Huang, Yang, Yuan, Su, Li, Su et~al.}]{zha2023tablegpt}
Liangyu Zha, Junlin Zhou, Liyao Li, Rui Wang, Qingyi Huang, Saisai Yang, Jing Yuan, Changbao Su, Xiang Li, Aofeng Su, et~al. 2023.
\newblock Tablegpt: Towards unifying tables, nature language and commands into one gpt.
\newblock \emph{arXiv preprint arXiv:2307.08674}.

\bibitem[{Zhang et~al.(2023{\natexlab{a}})Zhang, Yue, Li, and Sun}]{zhang2023tablellama}
Tianshu Zhang, Xiang Yue, Yifei Li, and Huan Sun. 2023{\natexlab{a}}.
\newblock Tablellama: Towards open large generalist models for tables.
\newblock \emph{arXiv preprint arXiv:2311.09206}.

\bibitem[{Zhang et~al.(2023{\natexlab{b}})Zhang, Henkel, Floratou, Cahoon, Deep, and Patel}]{zhang2023reactable}
Yunjia Zhang, Jordan Henkel, Avrilia Floratou, Joyce Cahoon, Shaleen Deep, and Jignesh~M Patel. 2023{\natexlab{b}}.
\newblock Reactable: enhancing react for table question answering.
\newblock \emph{arXiv preprint arXiv:2310.00815}.

\bibitem[{Zhao et~al.(2022)Zhao, Nan, Qi, Zhang, and Radev}]{zhao2022reastap}
Yilun Zhao, Linyong Nan, Zhenting Qi, Rui Zhang, and Dragomir Radev. 2022.
\newblock Reastap: Injecting table reasoning skills during pre-training via synthetic reasoning examples.
\newblock \emph{arXiv preprint arXiv:2210.12374}.

\end{thebibliography}

\clearpage
\appendix
\DoToC

\section{Ethics}
ChatGPT + Google Gemini 2.5 Pro was used for editing the manuscript for grammar and flow.

As we use publicly available datasets and models, there is a minimal chance of violating ethics guidelines. However, the use of large language models (LLMs) inherently raises some ethical concerns, particularly related to training data and environmental costs. 
Firstly, LLMs are trained on vast datasets that often include unfiltered internet data, which can contain biased, offensive, or otherwise harmful content. 
This can lead to the propagation of these biases in the model's outputs, affecting fairness and potentially causing harm when deployed in sensitive applications. 
Additionally, the lack of transparency regarding the sources and handling of this data can raise privacy issues, especially if personal or sensitive information is inadvertently included. 
Secondly, the environmental impact of training LLMs is considerable. 
Training these models requires substantial computational resources, leading to high energy consumption and a significant carbon footprint. 
This environmental cost is particularly concerning in the context of the ongoing climate crisis, as the energy used often comes from non-renewable sources. 
As such, there is a pressing need for the AI community to develop more sustainable and ethical practices, such as improving the efficiency of training processes and ensuring greater transparency and accountability in the use of training data.

\subsection{Reproducibility}
We obtained WikiTQ~\cite{pasupat2015compositional} \footnote{\url{https://github.com/ppasupat/WikiTableQuestions}}, TabFact~\cite{chen2019tabfact}\footnote{\url{https://github.com/wenhuchen/Table-Fact-Checking}}, and FeTaQA~\cite{nan2022fetaqa}\footnote{\url{https://github.com/Yale-LILY/FeTaQA}} from their GitHubs respectively. Evaluation was performed using Rouge score for FeTaQA and accuracy\footnote{\url{https://github.com/Leolty/tablellm}} for TabFact and WikiTQ. 

No updates to model weights were made. Inference was performed on the default training, validation, and test splits. Hyperparameters were kept as default. We used a temperature of 0.0 for consistency of our methods, and a temperature of 0.8 when running Mixed Self Consistency.

For the crossencoder, we ran inference on 500 samples of randomly-sampled validation data. We then trained the cross-encoder with the usefulness labels for 1 epoch at lr=2e-5.

Performing inference over the training+test set of WikiTQ and the validation+test set of Tabfact took around 2 weeks, with an estimated cost of OpenAI fees of \$3200. Ablations took around another week.

\subsection{Broader Impact}
Ideally, \method should be able to augment a real human Data Scientist in their everyday tasks in working with tabular data.
By effectively combining the strengths of both finetuning and prompting methods, \method can potentially make it more accessible and efficient for the average user.
This could lead to faster and more accurate data-driven decision-making in various industries, including healthcare, finance, and general research. 
Moreover, the cost-efficiency and interpretability of \method make it a viable option for organizations with limited computational resources or bias concerns, thereby democratizing access to advanced tabular reasoning capabilities. 
However, as with all LLMs, there is a potential for biased outcomes based on the training data, so careful, human supervision will always be needed at some step.

\section{Cost Analysis}
One of the significant advantages of \method is its computational efficiency \textbf{at inference time}. It is difficult to quantify the expense of \method training since these are entirely new steps that previous approaches do not use. Rather, previous approaches rely on their initial prompts, which may require significant tuning. One perspective would be that this shifts the computational burden. Still, at inference, this makes it an attractive option compared to other, more expensive approaches like Chain-of-Table, Mixed Self-Consistency, and H-STAR. These models often require 5 or more prompts to generate a predicted answer for a single table understanding task, leading to higher computational costs.

For instance, Chain-of-Table requires at least two separate prompts: one to generate intermediate reasoning steps and another to produce the final answer. H-Star requires 5 prompts at a minimum. \method, on the other hand, often only requires 1 pass, if it chooses not to use code (as seen in Table~\ref{tab:cost}). This increase in prompt length and the need for multiple stages of processing can substantially increase the model’s runtime and token usage. Similarly, approaches like Mixed Self-Consistency and DATER rely on multiple runs of self-consistency, where the model iterates over the same input multiple times to improve answer accuracy. This approach demands substantial computational resources, as each additional pass over the data incurs extra inference costs.

In contrast, \method leverages a more direct approach, allowing the model to dynamically choose whether to use code or direct reasoning in a single pass. This not only improves efficiency but also reduces the latency and cost per query, as it avoids the overhead of multiple inference calls. 
\begin{table}[ht]
\caption{Mean \# times code is called per test datapoint \label{tab:cost}}
\centering
\resizebox{.6\linewidth}{!}{
\begin{tabular}{lll} \toprule
 & GPT 4o-mini & GPT 4o \\ \midrule
WikiTQ & 1.105 & 0.910 \\
TabFact & 0.783 & 0.322 \\ \bottomrule
\end{tabular}
}
\end{table}
As shown in Table~\ref{tab:cost}, we see that most of the time, less than 1 call to code is needed. The stronger the model and the "easier" the task, the lower the number of calls is needed.

\section{Additional Baselines}
\label{app:addition_experiments}
In this section, we demonstrate results on methods that we were unable to run, but felt the need to include, to provide a full picture of our results. Table~\ref{tab:more_resuts} demonstrates these. We see that \method can outperform these baselines by a wide margin.

\begin{table}[ht]
\caption{More results. Note that these baselines are not trained, and results are taken from the best version from the original sources. \label{tab:more_resuts}}
\centering
\resizebox{\linewidth}{!}{
\begin{tabular}{l l l l} \toprule
\textbf{Approach}& \textbf{Base Model}& \textbf{WikiTQ}& \textbf{TabFact}\\ \midrule
BINDER& GPT-3 Codex& 64.60& 86.00\\
DATER& GPT-3 Codex& 65.90& 85.60\\
ReAcTable& code-davinci-002& 68.0 & 86.1\\
STRUCTGPT& GPT-3.5-turbo& 57.00& 87.60\\
ProTrix& GPT-3.5-turbo& 65.20& 83.50\\
Chain-of-Table& GPT-3.5-turbo& 59.94& 80.20\\
Chain-of-Table& PaLM 2& 67.31& 86.61\\ \midrule

\method\ -Reranker (Ours)& GPT-4o-mini& 75.84& 89.25\\
 \method (Ours)& GPT-4o-mini& 76.80&89.74\\
 \method\ -Reranker (Ours)& GPT 4o& 78.27&93.75\\
\method (Ours)& GPT 4o& 80.02& 93.38\\ \bottomrule
\end{tabular}
}
\end{table}

\section{Correction Statistics}
\label{app:sc_stats}

\begin{table}[ht]
\centering
\caption{Statistics on how many samples were correct and were able to be corrected by the prompt condition. Note that the number of prompt conditions are equivalent to the "Correct" columns. \label{tab:corrected}}
\resizebox{\linewidth}{!}{
\begin{tabular}{lrrrr} \toprule
 & \multicolumn{2}{c}{WikiTQ} & \multicolumn{2}{c}{TabFact} \\
 & Correct & Corrected & Correct & Corrected \\ \midrule
GPT 4o-mini & 2142 & 256 & 2550 & 151 \\
GPT 4o & 2232 & 317 & 2742 & 85 \\ \bottomrule
\end{tabular}
}
\end{table}

Table~\ref{tab:corrected} provides an analysis of the number of correct and corrected examples in the training set for WikiTableQA and TabFact, comparing GPT-4o-mini and GPT-4o. The results highlight the relationship between model capability and error correction, shedding light on how different models handle data quality.

Across both datasets, GPT-4o produces more correct answers than GPT-4o-mini, demonstrating its stronger reasoning and comprehension abilities. This gap is more pronounced in TabFact, where GPT-4o achieves a significantly higher number of correct responses, leaving fewer errors to be corrected. The relatively low number of corrections (85) suggests that GPT-4o already performs well on this dataset, reducing the need for additional intervention.
In contrast, WikiTableQA sees a higher number of corrections, particularly in GPT-4o-mini (256) and even more so in GPT-4o (317). 

\subsection{Qualitative Takeaways}
Qualitatively, the main categories of mistakes are either coding-related, logic-related, or formatting-related (e.g. missing an ampersand in the answer). 

Coding often introduced more logical errors--in one instance, the LLM filled the NANs with 0s, and then took a sum of the number of cells $<$ 4, which yielded the incorrect answer.

Coding often introduced so many errors that, if the retrieved prompt condition or the CoT suggested using code, the model would often fall into the trap of using Python in an example where they did not need to. Often, this resulted in lower performance. To fix this, we added the line \texttt{``Do not let these examples limit your creativity or decision to directly answer the question vs using code.''}. This generally solved the issue.

\section{Results on FeTaQA}
\label{app:fetaqa}

We choose to omit some results from FeTaQA from the main results given its inconsistent nature, aligning with previous work \cite{wang2024chain, abhyankar2024h}. Table~\ref{tab:FeTaQA Comparisons} shows our results. Like previous work, we found that the Rouge score\footnote{\url{https://github.com/huggingface/evaluate/blob/main/metrics/rouge/rouge.py}} evaluations missed many valid, but incorrectly formatted answers. We demonstrate examples in Table~\ref{tab:FeTaQA Bad}. 

\begin{table}[ht]
\centering
\caption{FeTaQA Comparisons \label{tab:FeTaQA Comparisons}.  (w/ PC) denotes With Prompt Condition retrieval}
\resizebox{\linewidth}{!}{
\begin{tabular}{llll} \toprule
Prompting& ROUGE-1& ROUGE-2& ROUGE-L\\ \midrule
End-to-End& 0.63& 0.41& 0.53\\
DATER& 0.63& 0.41& 0.53\\
Chain-of-Table& 0.66& 0.44& 0.56\\
H-STAR& 0.62& 0.39& 0.52\\ \midrule
\method\ (4o-mini) & 0.64& 0.41 & 0.53\\
\method\ -Reranker(4o-mini) & 0.65 & 0.40& 0.52\\ \bottomrule
\end{tabular}
}
\end{table}

\begin{table}[ht]
\centering
\caption{FeTaQA predictions that had ROUGE1 score $<$ 0.5 \label{tab:FeTaQA Bad}}
\resizebox{\linewidth}{!}{
\begin{tabular}{p{4cm}|p{4cm}} \toprule
Answer & Prediction \\ \midrule
after 25 total appearances for gamba osaka in the 2004 j1 league season miyamoto appeared 36 times in the 2005 season
&
tsuneyasu miyamotos total appearances increased from 25 in 2004 to 36 in 2005 \\ \midrule
alex yunevich served at central state from 1934 to 1936 and at alfred university from 1937 to 1976 compiling a career college football record of 186\textbackslash 98\textbackslash  13
&
alex yunevich achieved a total record of 186 wins 98 losses and 13 ties during his football career playing for the central state bearcats and the alfred saxons
\\ \midrule
the rg has a halflife of about 107 minutes and the rg has a halflife of about 51 minutes
&
the most recently published halflife measurements are 107 minutes for 286rg and 51 minutes for 283rg \\ \bottomrule
\end{tabular}
}
\end{table}

\section{Additional Figures}
\begin{figure}[ht]
    \centering
    \includegraphics[width=\linewidth]{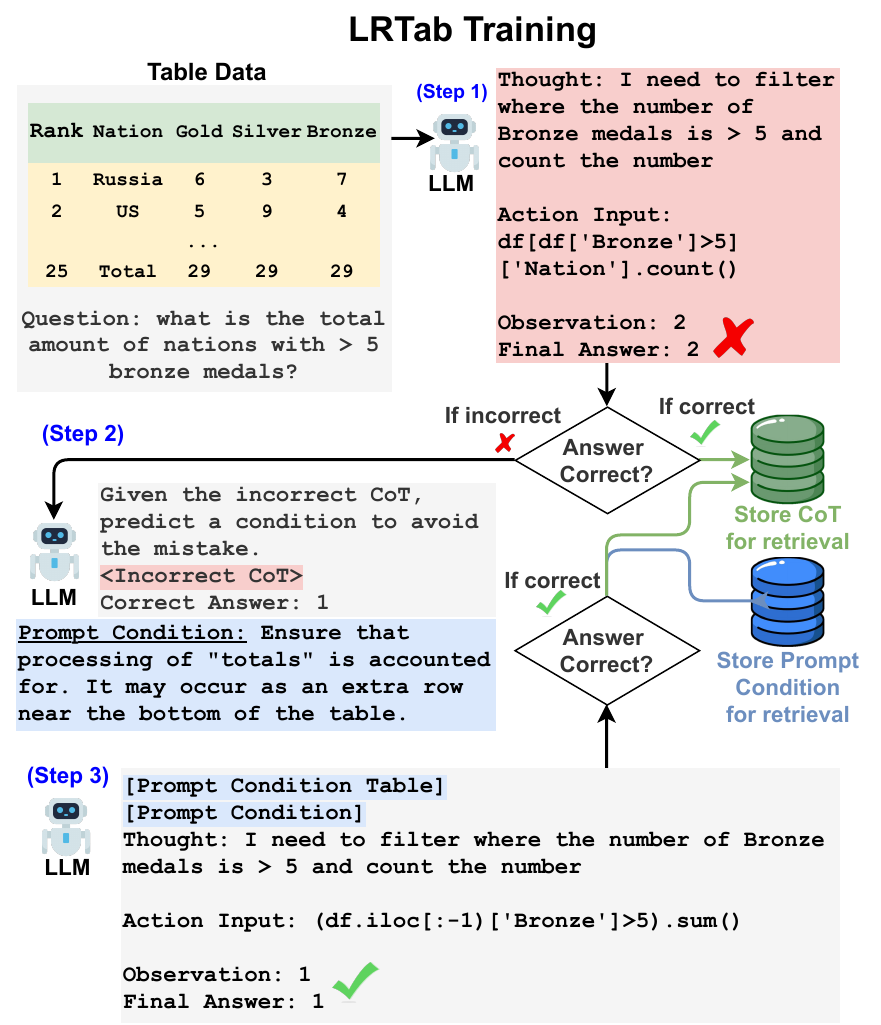}
    \caption{
    We obtain CoT Examples that are initially correct as well as those that were able to be corrected by the generated Prompt Conditions.
    \label{fig:ex_sc}}
\end{figure}

\clearpage
\section{Prompts}
\label{app:prompts}
\subsection{Examples of Prompts for WikiTQ}
Figure~\ref{fig:WikiTQ Direct Prompting} shows the Direct Prompting prompt used for the tabular question-answer task. 

Figure~\ref{fig:WikiTQ Code-Enabled Agent} is a prompt for the flexible, Code-Enabled Agent. It requires the LLM to solve the tabular question-answering task by optionally generating a Python program. If needed, we attempt to execute the resulting program to obtain the final answer. Note that for the experiments in Section~\ref{sec:optional_code}, we simply remove all mentions of optionality, and imply that code should always be used.

Figure~\ref{fig:WikiTQ Correction} shows the Correction Prompt proposed by our method. This prompt attempts to correct an incorrectly reasoned Previous CoT, and seeks to output a condition that would have prevented the error. This condition is added back into the previous prompt to check if it actually prevented the issue.
\subsection{Examples of Prompts for TabFact}
The following are the prompts used for \method on TabFact. The main difference from WikiTQ is the fact that the task is fact verification, so we inform the prompt there is a [STATMENT] to Verify as opposed to [QUESTION] to answer, and that the answer should only be True or False; otherwise, the prompt is the same.

The Direct Prompting prompt is shown in Figure~\ref{fig:TabFact Direct Prompting}, Code-Enabled Agent in Figure~\ref{fig:TabFact Code-Enabled Agent}, and Correction Prompt in Figure~\ref{fig:TabFact Correction}.

The main difference is the fact that the task is fact verification, so we inform the prompt there is a Statement to Verify as opposed to Question to answer, and that the answer should only be True or False.
\subsection{Examples of Prompts for FeTaQA}
The following are the prompts used for \method on FeTaQA. The main difference is that the questions follow a natural answer format. The correction prompt is also slightly adjusted to emphasize that the metric is based on n-gram similarity.

The Direct Prompting prompt is shown in Figure~\ref{fig:FeTaQA Direct Prompting}, Code-Enabled Agent in Figure~\ref{fig:FeTaQA Code-Enabled Agent}, and Correction Prompt in Figure~\ref{fig:FeTaQA Correction}.

\begin{figure*}
\begin{tcolorbox}
{\fontfamily{lmtt} \scriptsize
You are an advanced AI capable of analyzing and understanding information within tables. \\

Related example:

[EXAMPLE]

[CONDITION]

Read the table below regarding "[TITLE]".

[TABLE]

Question: [QUESTION]

Let's think step by step, and then give the final answer. Ensure the final answer is the following JSON format:\\

\`{}\`{}\`{}json

\{

"Final Answer": ... (AnswerName1, AnswerName2... form, no other form. And ensure the final answer is a number or entity names, as short as possible, without any explanation.)

\}

\`{}\`{}\`{}
}\end{tcolorbox}
\caption{WikiTQ Direct Prompting\label{fig:WikiTQ Direct Prompting}}
\end{figure*}

\begin{figure*}
\begin{tcolorbox}
{\fontfamily{lmtt} \scriptsize
You are working with a pandas dataframe in Python. The name of the dataframe is `df`. You may **optionally** use python code to transform the table to answer the question.\\
- All cells in the table should be considered as `object` data type, regardless of their appearance.

[CONDITION]

Related example:

[EXAMPLE]
\\\\
Strictly follow the given format to respond:\\

Thought: Reason step by step to answer the question. At the end, either directly provide the final answer or Action Input: \\

Action Input: Optional python pandas table transformation code. You must print out the output of the code.

Observation: Optional result of the action. If you get an error, just try to answer the question without fixing the error.

(Thought/Action Input/Observation can repeat, but use sparingly. Try to avoid repeating the same action.)\\

\`{}\`{}\`{}json

\{

"Final Answer": ... (AnswerName1, AnswerName2... form, no other form. And ensure the final answer is a number or entity names, as short as possible, without any explanation.)

\}

\`{}\`{}\`{}

You are provided with a table regarding "[TITLE]". This is the result of `print(df.to\_markdown())`:

[TABLE]

Question: [QUESTION]

Begin!

Thought:
}\end{tcolorbox}
\caption{WikiTQ Code-Enabled Agent \label{fig:WikiTQ Code-Enabled Agent}}
\end{figure*}

\begin{figure*}
\begin{tcolorbox}
{\fontfamily{lmtt} \scriptsize
You are an advanced AI capable of analyzing and understanding information within tables. Your task to provide a prompt condition that could have avoided the error in the previous COT. \\

You are provided with a table regarding "[TITLE]". 

[TABLE]

Question: [QUESTION] \\

Previous COT: 

You are given a previous, incorrect chain of thought for the table qa
Correct Answer: the correct answer to the table qa. Only use this for returning a useful prompt condition. Do not refer to this specific answer in the condition.\\

Previous COT:

[EXAMPLE]

Correct Answer: [ANSWER]\\

Reason step by step. At the end, return a prompt condition in JSON format shown below. 

\`{}\`{}\`{} json

\{

    "Condition": "..." (a prompt condition to prevent the mistake. For example: 'If the error occurred due to incorrect dataframe construction, try directly reasoning over the table instead of creating a dataframe.')

\}

\`{}\`{}\`{}
}\end{tcolorbox}
\caption{WikiTQ Correction \label{fig:WikiTQ Correction}}
\end{figure*}

\begin{figure*}
\begin{tcolorbox}
{\fontfamily{lmtt} \scriptsize
You are an advanced AI capable of analyzing and understanding information within tables. \\

Related example:

[EXAMPLE]

[CONDITION]

Read the table below regarding "[TITLE]".

[TABLE]

Based on the given table, check whether the following statement is true or false:

[STATEMENT]

Let's think step by step, and then give the final answer. Ensure the final answer is the following JSON format:\\

\`{}\`{}\`{}json

\{

"Final Answer": True/False (can only be True or False)

\}

\`{}\`{}\`{}
}\end{tcolorbox}
\caption{TabFact Direct Prompting\label{fig:TabFact Direct Prompting}}
\end{figure*}

\begin{figure*}
\begin{tcolorbox}
{\fontfamily{lmtt} \scriptsize
You are working with a pandas dataframe in Python. The name of the dataframe is `df`. You may **optionally** use python code to transform the table for the table fact-checking task.\\
- All cells in the table should be considered as `object` data type, regardless of their appearance.

[CONDITION]

Related example:

[EXAMPLE]
\\\\
Strictly follow the given format to respond:\\

Thought: Reason step by step to fact-check the statement. At the end, either directly provide the final answer or Action Input: \\

Action Input: Optional python pandas table transformation code. You must print out the output of the code.

Observation: Optional result of the action. If you get an error, just try to fact-check the statement without fixing the error.

(Thought/Action Input/Observation can repeat, but use sparingly. Try to avoid repeating the same action.)\\

\`{}\`{}\`{}json

\{

"Final Answer": True/False (can only be True or False)

\}

\`{}\`{}\`{}

You are provided with a table regarding "[TITLE]". This is the result of `print(df.to\_markdown())`:

[TABLE]

Based on the given table, check whether the following statement is true or false:

[STATEMENT]

Begin!

Thought:
}\end{tcolorbox}
\caption{TabFact Code-Enabled Agent\label{fig:TabFact Code-Enabled Agent}}
\end{figure*}

\begin{figure*}
\begin{tcolorbox}
{\fontfamily{lmtt} \scriptsize
You are an advanced AI capable of analyzing and understanding information within tables. Your task to provide a prompt condition that could have avoided the error in the previous COT. \\

You are provided with a table regarding "[TITLE]". 

[TABLE]

Statement to fact-check: [STATEMENT] \\

Previous COT: 

You are given a previous, incorrect chain of thought for the table fact-checking task

Correct Answer: the correct answer to the table qa. Only use this for returning a useful prompt condition. Do not refer to this specific answer in the condition.\\

Previous COT:

[EXAMPLE]

Correct Answer: [ANSWER]\\

Reason step by step. At the end, return a prompt condition in JSON format shown below. 

\`{}\`{}\`{} json

\{

    "Condition": "..." (a prompt condition to prevent the mistake. For example: 'If the error occurred due to incorrect dataframe construction, try directly reasoning over the table instead of creating a dataframe.')

\}

\`{}\`{}\`{}
}\end{tcolorbox}
\caption{TabFact Correction\label{fig:TabFact Correction}}
\end{figure*}

\begin{figure*}
    \begin{tcolorbox}
{\fontfamily{lmtt} \scriptsize
You are an advanced AI capable of analyzing and understanding information within tables. \\
- Example Response Style: 

    "Thompson prevailed in the 1982 Illinois gubernatorial election by a 5,074 vote margin."

    "In the 1874 New Jersey gubernatorial election, Joseph D. Bedle of the Democratic party defeated George A. Halsey of the Republican party with 53.65\% of the vote."\\

Related example:

[EXAMPLE]

[CONDITION]

Read the table below regarding "[TITLE]".

[TABLE]

Question: [QUESTION]

Let's think step by step, and then give the final answer. Ensure the final answer is the following JSON format:\\

\`{}\`{}\`{}json

\{

 "Final Answer": ... (Ensure the final answer format is a concise sentence similar to the Example Response Style.)

\}

\`{}\`{}\`{}
}\end{tcolorbox}
\caption{FeTaQA Direct Prompting\label{fig:FeTaQA Direct Prompting}}
\end{figure*}

\begin{figure*}
\begin{tcolorbox}
{\fontfamily{lmtt} \scriptsize
You are working with a pandas dataframe in Python. The name of the dataframe is `df`. You may **optionally** use python code to transform the table to answer the question.\\
- All cells in the table should be considered as `object` data type, regardless of their appearance.

- Example Response Style: 

    "Thompson prevailed in the 1982 Illinois gubernatorial election by a 5,074 vote margin."

    "In the 1874 New Jersey gubernatorial election, Joseph D. Bedle of the Democratic party defeated George A. Halsey of the Republican party with 53.65\% of the vote."\\

[CONDITION]

Related example:

[EXAMPLE]
\\\\
Strictly follow the given format to respond:\\

Thought: Reason step by step to answer the question. At the end, either directly provide the final answer or Action Input: \\

Action Input: Optional python pandas table transformation code. You must print out the output of the code.

Observation: Optional result of the action. If you get an error, just try to answer the question without fixing the error.

(Thought/Action Input/Observation can repeat, but use sparingly. Try to avoid repeating the same action.)\\

\`{}\`{}\`{}json

\{

 "Final Answer": ... (Ensure the final answer format is a concise sentence similar to the Example Response Style.)

\}

\`{}\`{}\`{}

You are provided with a table regarding "[TITLE]". This is the result of `print(df.to\_markdown())`:

[TABLE]

Question: [QUESTION]

Begin!

Thought:
}\end{tcolorbox}
\caption{FeTaQA Code-Enabled Agent\label{fig:FeTaQA Code-Enabled Agent}}
\end{figure*}

\begin{figure*}
\begin{tcolorbox}
{\fontfamily{lmtt} \scriptsize
You are an advanced AI capable of analyzing and understanding information within tables. Your task to provide a prompt condition to avoid the error in the previous COT and rephrase the answer to match the Correct Answer in terms of n-gram overlap. \\

You are provided with a table regarding "[TITLE]". 

[TABLE]

Question: [QUESTION] \\

Previous COT: 

You are given a previous, incorrect chain of thought for the table qa
Correct Answer: the correct answer to the table qa. Only use this for returning a useful prompt condition. Do not refer to this specific answer in the condition.\\

Previous COT:

[EXAMPLE]

Correct Answer: [ANSWER]\\

Reason step by step. At the end, return a prompt condition in JSON format shown below. 

\`{}\`{}\`{} json

\{

    "Condition": "..." (a prompt condition that could have avoided the error in the previous COT and rephrase the answer to match the Correct Answer. For example: 'If the error occurred due to incorrect dataframe construction, try directly reasoning over the table instead of creating a dataframe. The format should be similar to [prepositional phrase, a main clause, and modifiers]')

\}

\`{}\`{}\`{}
}\end{tcolorbox}
\caption{FeTaQA Correction\label{fig:FeTaQA Correction}}
\end{figure*}

\section{Full Example}
Figure~\ref{fig:full_example} shows a complete example of \method with both Prompt Conditions and CoT Retrieval.

\begin{figure*}[ht]
\begin{tcolorbox}
{\fontfamily{lmtt} \scriptsize
You are working with a pandas dataframe in Python. The name of the dataframe is `df`. You may **optionally** use python code to transform the table to answer the question.

- All cells in the table should be considered as `object` data type, regardless of their appearance.

Potentially relevant examples:

Example Table:

|    | Team 1   | Agg.   | Team 2              | 1st leg   | 2nd leg   |

|---:|:---------|:-------|:--------------------|:----------|:----------|

|  0 | Benfica  | 3–2    | Paris Saint-Germain | 2–1       | 1–1       |

...

|  7 | Braga    | 1–08   | Liverpool           | 1–0       | 0–0       |

Example Question: how many total points did the winning teams score in the round of 16?

What to watch out for: Ensure that all data entries are correctly formatted  and validated before performing calculations. Specifically, check for and  correct any inconsistencies in score formatting, such as extra characters or incorrect delimiters, to accurately determine the winning team and their total goals.

Example Table:

|    |   Rank | Bib                 | Name           | Nationality   | Start       | Penalties (P+P+S+S)   |   Time |   Deficit |

|---:|-------:|:--------------------|:---------------|:--------------|:------------|:----------------------|-------:|----------:|

|  0 |      1 | Emil Hegle Svendsen | Norway         | 0:00          | 1 (0+0+0+1) | 32:35.5               |    nan |       nan |

...

| 59 |     52 | Michal Slesingr     | Czech Republic | 2:18          | nan         | DNS                   |    nan |       nan |

Example Question: how many took at least 35:00 to finish?

What to watch out for: Ensure that the 'Time' column is treated as strings and compared lexicographically when filtering for specific time thresholds, especially when the format is non-standard like 'mm:ss.s'.

Use them to understand the context. Do not let the examples limit your creativity.

Related example:

|    | Pos   | Rider             | Manufacturer   | Time/Retired   |   Points |

|---:|:------|:------------------|:---------------|:---------------|---------:|

|  0 | 1     | Max Biaggi        | Aprilia        | 40:36.299      |       25 |

...

| 30 | Ret   | Takeshi Tsujimura | Honda          | Retirement     |      nan |

To find the difference between Marcellino Lucchi's points and Max Biaggi's points, I need to locate their respective rows in the dataframe and subtract their points.

1. Max Biaggi is in the first row with 25 points.

2. Marcellino Lucchi is in the second row with 20 points.

The difference in their points is (25 - 20 = 5).

\`{}\`{}\`{}json
\{
 "Final Answer": 5
\}
\`{}\`{}\`{}

Do not let the example limit your creativity or decision to directly answer the question vs using code.

Strictly follow the given format to respond:

Thought: Reason step by step to answer the question. At the end, either directly provide the final answer or Action Input:

Action Input: Optional python pandas table transformation code. You must print out the output of the code.

Observation: Optional result of the action. If you get an error, just try to answer the question without fixing the error.

(Thought/Action Input/Observation can repeat, but use sparingly. Try to avoid repeating the same action.)

\`{}\`{}\`{}json
\{
 "Final Answer": ... (AnswerName1, AnswerName2... form, no other form. Ensure the final answer is a number or entity names, as short as possible, without any explanation.)
\}
\`{}\`{}\`{}

You are provided with a table regarding "2008 Clasica de San Sebastian". This is the result of `print(df.to\_markdown())`:

|    |   Rank | Cyclist                  | Team               | Time       |   UCI ProTour Points | Nationality   |

|---:|-------:|:-------------------------|:-------------------|:-----------|---------------------:|:--------------|

|  0 |      1 | Alejandro Valverde (ESP) | Caisse d'Epargne   | 5h 29' 10" |                   40 | ESP           |

|  1 |      2 | Alexandr Kolobnev (RUS)  | Team CSC Saxo Bank | s.t.       |                   30 | RUS           |

|  2 |      3 | Davide Rebellin (ITA)    | Gerolsteiner       | s.t.       |                   25 | ITA           |

|  3 |      4 | Paolo Bettini (ITA)      | Quick Step         | s.t.       |                   20 | ITA           |

|  4 |      5 | Franco Pellizotti (ITA)  | Liquigas           | s.t.       |                   15 | ITA           |

|  5 |      6 | Denis Menchov (RUS)      | Rabobank           | s.t.       |                   11 | RUS           |

|  6 |      7 | Samuel Sanchez (ESP)     | Euskaltel-Euskadi  | s.t.       |                    7 | ESP           |

|  7 |      8 | Stephane Goubert (FRA)   | Ag2r-La Mondiale   | + 2"       |                    5 | FRA           |

|  8 |      9 | Haimar Zubeldia (ESP)    | Euskaltel-Euskadi  | + 2"       |                    3 | ESP           |

|  9 |     10 | David Moncoutie (FRA)    | Cofidis            | + 2"       |                    1 | FRA           |

Question: how many more points did team liquigas score than robobank?

Begin!

Thought:

To determine how many more points Team Liquigas scored than Rabobank, I need to locate the rows corresponding to these teams and compare their UCI ProTour Points.

1. Find the row for Team Liquigas and note the UCI ProTour Points.

2. Find the row for Rabobank and note the UCI ProTour Points.

3. Subtract the points of Rabobank from those of Liquigas to find the difference.

Action Input: 
\`{}\`{}\`{}python

\# Extracting the points for Team Liquigas and Rabobank

liquigas\_points = df[df['Team'] == 'Liquigas']['UCI ProTour Points'].values[0]

rabobank\_points = df[df['Team'] == 'Rabobank']['UCI ProTour Points'].values[0]

\# Calculating the difference

difference = liquigas\_points - rabobank\_points

print(difference)
\`{}\`{}\`{}

Observation: 
4

\`{}\`{}\`{}json
\{
 "Final Answer": 4
\}
\`{}\`{}\`{}
}\end{tcolorbox}
\caption{Full Example\label{fig:full_example}}
\end{figure*}

\end{document}